\pdfoutput=1
\documentclass[11pt]{article}

\usepackage[final]{acl}

\usepackage{times}
\usepackage{color,soul}
\usepackage{latexsym}
\usepackage{graphicx}% so it makes black blobs
\usepackage{xcolor,lipsum,subcaption}
\usepackage{inconsolata}
\usepackage{svg}
\usepackage{makecell}
\usepackage{xcolor, soul}
\usepackage{soulpos}
\DeclareRobustCommand{\hlcyan}[1]{{\sethlcolor{cyan!20}\hl{#1}}}
\usepackage[T1]{fontenc}
\usepackage[utf8]{inputenc}

\usepackage{microtype}
\usepackage{hyperref}
\usepackage{inconsolata}

\title{DeepPavlov at SemEval-2024 Task 8: Leveraging Transfer Learning for Detecting Boundaries of Machine-Generated Texts}

\author{Anastasia Voznyuk \and Vasily Konovalov \\
 Moscow Institute of Physics and Technology\\
  \texttt{\{vozniuk.ae, vasily.konovalov\}@phystech.edu}\\
 }

\begin{document}
\maketitle
\begin{abstract}
The Multigenerator, Multidomain, and Multilingual Black-Box Machine-Generated Text Detection shared task in the SemEval-2024 competition aims to tackle the problem of misusing collaborative human-AI writing. Although there are a lot of existing detectors of AI content, they are often designed to give a binary answer and thus may not be suitable for more nuanced problem of finding the boundaries between human-written and machine-generated texts, while hybrid human-AI writing becomes more and more popular. In this paper, we address the boundary detection problem. Particularly, we present a pipeline for augmenting data for supervised fine-tuning of DeBERTaV3. We receive new best MAE score, according to the leaderboard of the competition, with this pipeline.
\end{abstract}
\section{Introduction}

Recently, there has been a rapid development of auto-regressive language models, for example, GPT-3~\cite{gpt3}, GPT-4~\cite{openai2023gpt4}, and LLaMA2~\cite{touvron2023llama2}. These models are trained on enormous amounts of data and are able to produce coherent texts that can be indistinguishable from human-written texts~\cite{dugan2022real}.

The SemEval-2024 Task~8 competition suggests to tackle the problem of detecting machine-generated texts. This problem has become more relevant recently due to the release of ChatGPT\footnote{\url{https://openai.com/blog/chatgpt}}, a model by OpenAI that simplified the access to the large language models (LLM) and their usage. For example, LLM can be maliciously used to generate fake news~\cite{zellers2020defending}. There are also some concerns raised among scientists~\cite{ma2023ai} and educators~\cite{zeng2023automatic} that the usage of LLMs will devalue the process of learning and research.

The commonly used approach to formulate the task of detecting machine-generated texts is a binary classification task~\cite{jawahar-etal-2020-automatic}. In this case, a text can be attributed to  either a human or a LLM. Otherwise, the task can be formulated as a multiclass classification or an authorship attribution task~\cite{uchendu-etal-2020-authorship}, where it is needed to determine which one of the $k$ authors is the real author of the given text.
Finally, the trend toward human-AI collaborative writing is rising, which highlights the importance of the boundary detection task. In this setup, text contains consecutive chunks of different authorship, and it is required to detect where the boundaries between chunks lie and who is the author of every chunk. Due to its complexity, it is usually assumed the text has a human-written prefix and the rest of the text is AI-generated~\cite{dugan2022real, Cutler2021AutomaticDO, kushnareva2023artificial}. \\
Our main contributions are three-fold:
\begin{enumerate}
    \item  We receive the new best MAE score on the task of detecting the boundary between human-written and machine-generated parts of the text.
    \item  We present a new simple yet effective pipeline of augmenting data for the task of boundary detection, which allows us to get more data for training and improve the results of fine-tuning large language models.
    \item We compare the performance of several fine-tuned models with different architectures on various amounts of training data.
\end{enumerate}

Additionally, we've made the code of augmentation publicly  available.\footnote{\url{https://github.com/natriistorm/semeval2024-boundary-detection}}

\section{Related Work}

Most of approaches~\cite{jawahar-etal-2020-automatic} for machine-generated text detection are based on calculating linguistic~\cite{Frhling2021FeaturebasedDO}, stylometric, and statistical features, as well as on using classical machine learning methods like logistic regression, random forest, and gradient boosting as classifiers. Among commonly used features are word and n-gram frequencies~\cite{manjavacas-etal-2017-assessing}, and tf-idf~\cite{solaiman2019release}.

An alternative strategy is to use zero-shot techniques based on the internal metrics of the texts. For example, token-wise log probability can be evaluated by models like GROVER~\cite{zellers2020defending} or GPT-2~\cite{solaiman2019release}. A probability threshold is established to distinguish writings produced by machines from those written by humans. Moreover, rank~\cite{gltr} or log-rank~\cite{mitchell2023detectgpt} can be calculated for each token and compared for consistency with the prior context.

It's shown by~\citet{ippolito-etal-2020-automatic} that feature-based methods are inferior to methods based on using encoders of pretrained language models like BERT~\cite{devlin-etal-2019-bert} as a basis for fine-tuning on the selected domain. Representations from auto-regressive language models can be used as input for the classification head. Such transformer-based methods require supervised detection examples for further training. Among the models commonly used for fine-tuning are RoBERTa~\cite{roberta} and XLNet~\cite{xlnet}.
Fine-tuned RoBERTa has shown the state-of-the-art results on the problems of authorship attribution~\cite{uchendu2021turingbench}.

To tackle the problem of mixed human-machine writing,~\citet{dugan2022real} introduces the Real Or Fake Text (RoFT) tool, where humans were asked to detect the sentence where the text transitions from human-written text to machine-generated text. One of the possible formulations of this task is multilabel classification~\cite{Cutler2021AutomaticDO}, where the boundary detector needs to determine the first generated sentence, and the number of this sentence is considered the label of the text. In that work, each sentence is processed separately with shallow classification and regression models based on RoBERTa and SRoBERTa~\cite{reimers2019sentencebert}. That solution perform well in an in-domain setup, but is limited in an out-of-domain setup. 

Any solution for tasks about detecting machine-generated texts should be robust to domain change. 
The organisers of the SemEval-2024 Task~8 competition claimed to have added new domains to the test set for testing the robustness of participants' solutions. There are several works on performance of detecting methods on out-of-domain setup, such as~\citet{kushnareva2023artificial} and~\citet{zeng2023automatic}. \citet{kushnareva2023artificial} conclude that perplexity-based and topological features appear to be helpful in case of domain shift.

 \section{Data and Task Description}
 \label{dataset_desc}
 \subsection{Task Description}

 The Multigenerator, Multidomain, and Multilingual Black-Box Machine-Generated Text Detection~\cite{semeval2024task8} is focused on challenging detectors of machine-generated texts.
    The dataset, provided by organisers, consists of 3 parts:
 \begin{enumerate}
     \item texts of different authorship for subtask A and subtask B;
     \item texts with collaborative human-AI writing for subtask C.
 \end{enumerate}
This paper focuses only on subtask C, suggesting a solution to differentiate a human-written prefix from the rest of the AI-generated text.  The texts for this subtask are generated in the following way: the language model should continue the human-written text, which is given as a prompt. Several examples of texts are presented in Appendix~\ref{appendix:examples}. The designated evaluation metric for this task is Mean Absolute Error (MAE), which quantifies the absolute distance between the predicted word and the actual word where the switch of authorship between human and LLM occurs.
 
 \subsection{Data Description}
 \begin{figure*}[!htb]
    \centering
    \subfloat[Distribution of lengths of tokenized texts]{%
        \includegraphics[width=0.5\linewidth]{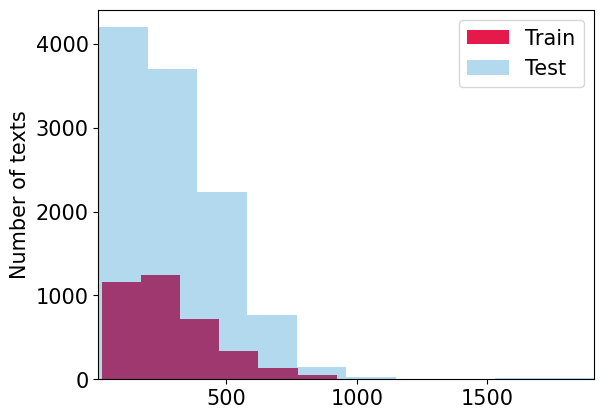}%
        \label{fig:a}%
        }%
    \hfill%
    \subfloat[Distribution of boundary subtoken positions]{%
        \includegraphics[width=0.5\linewidth]{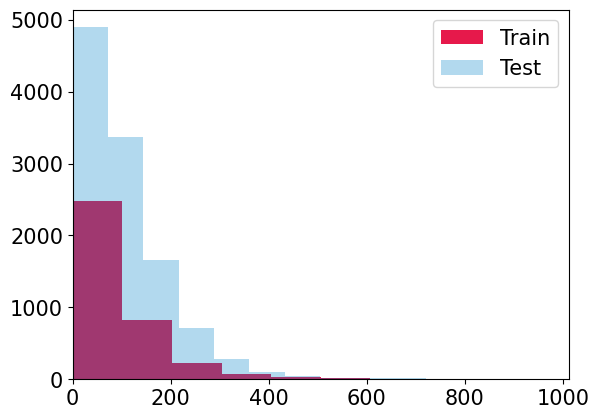}%
        \label{fig:b}%
        }%
    \caption{Statistics of the texts in the datasets}
\end{figure*}

The dataset for this task is derived from the M4~\cite{wang2023m4} dataset, which contains  texts of various domains, various languages, and generators. The authors show that current detectors tend to misclassify machine-generated texts as human-written if they're given a text from a different domain. The texts in the train and validation datasets are generated from scientific paper reviews from PeerRead~\cite{kang-etal-2018-dataset}. The test set partially consists of texts generated from PeerRead.  In order to check the robustness of the solutions to domain shift, texts from Outfox, a dataset of LLM-generated student essays~\cite{koike2023outfox}, are added to the dataset. All data sets contain English texts only.

 \subsection{Data Analysis} 

The distribution of lengths of texts in train and test data sets, tokenized by DeBERTa-V3-base models, is shown in Figure~\ref{fig:a}.
The majority of the texts in the test set are shorter in length, but there are several texts that exceed 1,000 subtokens.
The distribution of positions of a boundary subtoken in texts is shown in Figure~\ref{fig:b}. A boundary subtoken is the first subtoken of a boundary word. In both datasets, there is a distinguishable disproportion in the position of a boundary subtoken, as most of them have the boundary subtoken within the first 200 subtokens. Thus, a model trained only  on this data will give limited results when encountering longer texts with longer human prefixes in them.

\section{Proposed Method}

The provided train dataset consists only of 3,649 texts. It's well known that an abundance of in-domain training data is crucial for classifier performance~\cite{konovalov2016collecting}. However, during the competition period, it was prohibited to use any external data for training, and thus we were limited to working only with the provided dataset. In this case, getting more training data with some kind of augmentation plays a crucial role. We designed an augmentation pipeline and ran all our experiments on two sets of data: the one provided by the organisers of the competition, described in Section~\ref{dataset_desc}, and our augmented data.

\subsection{Data Augmentation}
  \label{sec:augment}
   The general idea of our augmentation pipeline is to split the text into distinct sentences and take several consequent sentences from the text with authorship change. It will make new texts coherent and will not mislead models during training~\cite{ostyakova-etal-2023-chatgpt}. In addition, to be useful for training, each sequence should contain a sentence with an authorship change.

    Another nuance about the augmentation process is the need to correctly determine the boundary label. The boundary label is calculated as a number of whitespace-split words in the human-written prefix. However, the initial dataset contains texts where a pair or even a sequence of words is split only by line breaks or punctuation symbols. Such a sequence of words should be considered as one word when calculating the boundary label. Thus, we have to pay a lot of attention to whitespace characters during augmentation and do not mistakenly append new whitespace characters between words.

    We preprocess each text in the dataset for augmentation in the following way: 
    \begin{enumerate}
        \item Split the text into sentences by punctuation symbols.
        \item Split the sentences themselves by whitespaces into lists of whitespace-split words.
        \item Compare the list of whitespace-split words from previous step with the list of whitespace-split words obtained by splitting the text itself. In case of discrepancy, fix it, depending on its type.
    \end{enumerate}
     See example of preprocessing for the text from train set in Appendix~\ref{appendix:pipeline}.
     
    In the third step of preprocessing, there are two main types of discrepancies: lost whitespace characters and words sequences, originally separated by line breaks only that were split during sentence split process. The former is solved by inserting the missing whitespace characters, while for the latter we merge the split words into one sequence.
    The last step of preprocessing is crucial and skipping it will result in the incorrect calculation of the boundary label for augmented text, as the label directly depends on the number of whitespace characters in the text. 
   
    After preprocessing, we take a number of consecutive sentences to the left and to the right from the boundary sentence, combine them in a text and determine the label of the boundary word in this new text.
    
\subsection{Model Comparison}
\label{sec:model_comparison}
We used only the transfer learning approach, where a pretrained transformer-based model is fine-tuned on our task in a supervised way. We would like the model to be able to work with long enough sequences because we want the whole human-written prefix to fit in the encoder. Thus, we've determined three models that showed good results on the task of machine-generated text detection:
    
    \begin{enumerate}
        \item RoBERTa~\cite{roberta} has shown good performance in both tasks of boundary  detection~\cite{kushnareva2023artificial} and  machine-generated text detection~\cite{macko-etal-2023-multitude}.
        \item Longformer~\cite{longformer}, which was suggested as a baseline by organisers of the task. This model is based on pretrained RoBERTa with novel attention mechanism with a sliding window to long sequences.
        \item DeBERTa~\cite{he2021deberta}, which is the state-of-the-art model for machine-generated text detection~\cite{macko-etal-2023-multitude}. It overcomes the BERT and RoBERTa  by introducing a disentangled attention for encoding the position and content of each token separately into two vectors. We decided to test it in the boundary detection task to understand whether it outperform RoBERTa. In our experiments, we fine-tuned DeBERTaV3~\cite{he2021debertav3} which is an enhanced version of DeBERTa.
\end{enumerate}

All three models are fine-tuned for token classification task. For each token, models predict the probability of being a boundary token and output the most probable token.

\section{Experimental Setup} 
       We have used pretrained \texttt{longformer-4096-base} and \texttt{longformer-4096-large} with default hyperparameters to fine-tune Longformer. 
        For experiments with RoBERTa, we have chosen two models: \texttt{roberta-base} and  \texttt{roberta-large}. The models were fine-tuned with the set of custom hyperparameters, taken from the original paper~\cite{roberta}. 
        Finally, for experiments with DeBERTa, we have also chosen two models:  \texttt{deberta-v3-base} and  \texttt{deberta-v3-large}. The models were fine-tuned with the set of custom hyperparameters taken from~\citet{he2021debertav3}. All custom hyperparameters are listed in Appendix~\ref{appendix:hyperparameters}.
        
        For all models we set the maximum length of context in tokenizer equal to 512 as there are only few text items in both train and test set with tokenized text length greater than 512. Additionaly, we've used the early stopping method for all of our experiments to get rid of epochs dependency.
        
~ 

\begin{table}[t!]
\centering
\begin{tabular}{lcc}
\textbf{Model} & \textbf{dev} & \textbf{test} \\
\hline
RoBERTa-base & 9.04 \textbackslash\hspace{0.25em}5.78 & 31.56 \textbackslash\hspace{0.25em}30.71\\
RoBERTa-large & 6.72 \textbackslash\hspace{0.25em}4.18 & 25.25 \textbackslash\hspace{0.25em}20.66\\
\hline
longformer-base  & 5.10 \textbackslash\hspace{0.25em}5.67 & 23.16 \textbackslash\hspace{0.25em}22.94 \\
longformer-large  & 4.54 \textbackslash\hspace{0.25em}4.40 & 22.97 \textbackslash\hspace{0.25em}20.33 \\
\hline
DeBERTaV3-base & 3.66 \textbackslash\hspace{0.25em}3.15  & 16.12 \textbackslash\hspace{0.25em}13.98 \\
DeBERTaV3-large & 2.38 \textbackslash\hspace{0.25em}2.54 & 15.16 \textbackslash\hspace{0.25em}\textbf{13.38} \\
\hline
Top-1 Submission & - & 15.68
\end{tabular}
\caption{MAE on original \textbackslash\hspace{0.25em}augmented dataset and comparison with Top-1 submission on the leaderboard. Longformer-base is suggested as a baseline solution by organisers of competition.} 
\label{tab:mae}
\end{table}

\section{Results and Discussion}
\subsection{Main Results}
\label{sec:results}

In Tables~\ref{tab:mae}, we compare MAE scores of different models from Section~\ref{sec:model_comparison}. There are experiments with models fine-tuned on the original dataset provided by organisers and on the  extended dataset with both augmented and original texts.

All models perform better when they are fine-tuned on the extended dataset rather than only on original texts. It clearly shows that even such a simple data augmentation provides a significant boost in performance.

If we compare the results among the models, we will clearly see the dominance of DeBERTaV3 models.
For both setups DeBERTaV3-large has shown the best performance and the lowest MAE score on the validation and test datasets. On the setup with the extended dataset, DeBERTaV3-large gets new best MAE score for the competition, which is equal to 13.375. It improves MAE score of the top-1 submission by more than 2 points, from 15.683 to 13.38.
%  that  demonstrates great performance on downstream tasks
\subsection{Discussion}
Results in Section~\ref{sec:results} show the importance of both the variety of the data in the dataset and a pretrained model. Leveraging  augmented training data significantly increases performance on the task, because it introduces variety in lengths of texts and boundary token positions while preserving the coherence of the texts. We believe that to be the reason why the models perform better when they are fine-tuned on the dataset with augmented data rather than on the original dataset.

Apart from various data, it is also important for the pretrained model to have great generalization capabilities. DeBERTaV3-large has better generalization capabilities than other tested models~\cite{he2021deberta}. The advanced architecture of DeBERTaV3 helps to significantly improve the MAE score in comparison with both RoBERTa and Longformer.

For all three models, large vesion of each model outperforms base version on both the original and the extended dataset. The reason to this is, greater number of trainable parameters allows models to generalize on training set better.

\subsection{Error Analysis}

 We manually inspected texts from the test set on which the best-performing model, DeBERTaV3-large, made serious mistakes. We've limited our inspection set to the texts where the distance between the predicted label and the true label was more than 100 tokens. Thus, we got 276 texts. Only 75 out of these 276 texts were from PeerReview domain, so model made most of its mistakes on the Outfox domain, texts from which are not present in the train set.
 
 These two domains presented in the test set are very different: they vary in the style of formatting, punctuation, and text structuring. The second domain of LLM-generated student essays have a lot of spelling and punctuation problems and it may confuse the models, as they trained on more formal and literate texts. It would be interesting to evaluate models on each of these domains separately. However, because we do not have domain classification lables in the test set, it is not yet feasible.

\subsection{Anomalies in Texts}

In the majority of texts from the original dataset, the model generates a coherent continuation of the human-written prefix, and it may be hard for a human to guess the boundary word without knowing it. However, there are a number of texts in the data sets that have some flaws in the generated parts.

There are texts in which LLM hallucinated. It either repeated the human prefix or went into a loop where it generated excessive lists with the same beginning. See example of it in Appendix~\ref{appendix:examples} and in Appendix~\ref{appendix:errors}. Such hallucinations can be an immediate hint for detector model to put the label boundary near this anomaly.
Also, sometimes machine-generated text can have some distinguished features that imply the artificial nature of a particular part of the text. For example, in a number of cases the model begins the generation with the """ (three double quotation marks). It may also be a hint for the detector.
A list of other common features we've encountered while examining the test set is provided in Appendix~\ref{appendix:errors}. 

\section{Conclusion}
In this paper we describe the system submitted for SemEval2024-Task~8, the subtask dedicated to hybrid human-machine writing detection. We present a simple yet effective augmentation pipeline. We explore how adding this pipeline to the process of fine-tuning can significantly increase the performance on the task, and provide an analysis of performance of various models with and without our augmentation pipeline. The best model, which is DeBERTaV3-large fine-tuned on a large set of augmented data, receives a new best score according to the leaderboard of the competition. Other fine-tuned models achieve competitive results, ranking in the upper half of the leaderboard and beating the organisers' baseline. As the provided data was limited to English language only, future work might include training multilingual boundary detection solution by mixing training data of different languages and using a multilingual encoder~\cite{chizhikova2022multilingual}. Such a system can be used for hybrid AI-writing detection as a standalone solution or can be integrated into existing NLP frameworks like DeepPavlov~\cite{burtsev2018deeppavlov}.

%\bibliographyf
\bibliography{custom}
\appendix

\section{Examples of Texts}
\label{appendix:examples}

Table~\ref{tab:examples} contains three examples of how authorship change occurs in the texts from the train set. While the first text contains no signs of a flawed generation, the second and third texts have some flaws. In the second text, the model begins to generate from the capital letter, and its generation is incoherent with the human-written prefix. In the third text, the model starts to repeat the end of the human part, which is a glaring sign of machine-generated text.

\begin{table}[ht!]
\centering
\begin{tabular}{l}
\hline
\textbf{Model}\\
\hline
\makecell{\hlcyan{I noticed that in Figure 2, the two quantization} \\ \hlcyan{factors for} quantized layers are missing labels.  \\  It would be helpful for the reader to understand \\  which layers are being quantized in the figure}. \\
\hline
\makecell{\hlcyan{Hi Authors, You seem to have submitted two} \\
\hlcyan{of the same paper? Pls advise} Could you please \\ clarify if this  is a mistake or if there are any \\  differences between the two submitted papers?} \\
\hline
\makecell{\hlcyan{There has been prior work on semi-supervised} \\ \hlcyan{GAN, though this paper is the first context} \\ \hlcyan{conditional variant. The novelty of the approach} \\  \hlcyan{was}
the novelty of the approach was  leveraging \\  in-painting using an adversarial loss to  gene- \\ rate contextually relevant images. } \\
\hline
\end{tabular}
\caption{Examples of texts from train set with different quality of LLM generation and with highlighted human prefix}
\label{tab:examples}
\end{table}

\section{Hyperparameters}
\label{appendix:hyperparameters}

For fine-tuning DeBERTaV3 we use hyperparameters, listed by model authors in \citet{he2021debertav3}. Table~\ref{tab:hyperparameters_large} lists these hyperparameters.
\begin{table}[!htb]
\centering
\begin{tabular}{lcc}
\hline
\textbf{Hyperparameters} & \textbf{Large}\ &  \textbf{Base}\\
\hline
Optimizer & AdamW & AdamW\\
Adam $\beta_1$, $\beta_2$ & 0.9, 0.999 & 0.9, 0.999\\ 
Adam $\epsilon$ & 1e-6 & 1e-6 \\ 
Warm-up step & 50 & 50\\
Batch size & 4 & 32 \\
Learning rate (LR) & 5e-6 & 2e-6 \\
Learning rate decay & Linear & Linear \\
Weight decay & 0.01 & 0.01 \\
Gradient clipping & 1.0 & 1.0 \\
\hline
\end{tabular}
\caption{Hyperparameters for fine-tuning DebertaV3-large and DebertaV3-base}
\label{tab:hyperparameters_large}
\end{table}

Table~\ref{tab:hyperparameters_roberta} contains hyperparameters for fine-tuning RoBERTa, also taken from original paper~\cite{roberta}.
\begin{table}[!htb]
\centering
\begin{tabular}{lc}
\hline
\textbf{Hyperparameters} & \textbf{values}\\
\hline
Optimizer & AdamW \\
Warm-up steps & 50 \\
Batch size & 16 \\
Learning rate (LR) & 5e-6 \\
Learning rate decay & Linear \\
Weight decay & 0.01 \\
\hline
\end{tabular}
\caption{Hyperparameters for fine-tuning RoBERTa}
\label{tab:hyperparameters_roberta}
\end{table}

\section{Examples For Error Analysis}
\label{appendix:errors}
See Table~\ref{tab:anomalies}.

\begin{table*}[ht!]
\centering
\begin{tabular}{lcc}
\textbf{Anomaly} & \textbf{Text Id} & \textbf{Example}\\
\hline
Excessive repetition & 613 & \makecell{...\hlcyan{praying for.} No more traffic jams, no more \\ parking nightmares, no more car payments, no \\ more insurance, no more maintenance, no more oil \\ changes, no more tire rotations...} \\
\hline
% Change from formal style to informal \\
% \hline
Extremely long lists of items & 8854 & \makecell{...\hlcyan{nice approach to} """\\
+ learning skills\\
+ learning skills in a sample efficient way\\
+ learning skills in an interpretable way \\
+ learning skills that can be used on downstream tasks \\
+ learning skills that are transferable between domains}\\
\hline
\makecell{JSON-structured hallucinations} & 5358 & \makecell{...\hlcyan{Summary of revisions:} \\
$\ast \ast \ast$ """,\\
    "title": "Diet Networks: Thin Parameters for \\ Fat Genomics",\\
    "abstract": "Learning tasks such as...} \\
%Excessive usage of adjectives\\
\hline
Bizarre formatting & 5639 & \makecell{...\hlcyan{Below are my comments:} \\\hlcyan{(1)} """ \\
The first sentence of the abstract is too \\ long. It should be divided into two sentences\\
(2) \\
"""} \\
\hline
\end{tabular}
\caption{Table with some frequent anomalies in the generated part of texts from test set. The highlighted part is human-written prefix.} 
\label{tab:anomalies}
\end{table*}

\section{Augmenting Pipeline Scheme}
\begin{figure*}[t!]
\centerline{\includegraphics[width=\linewidth]{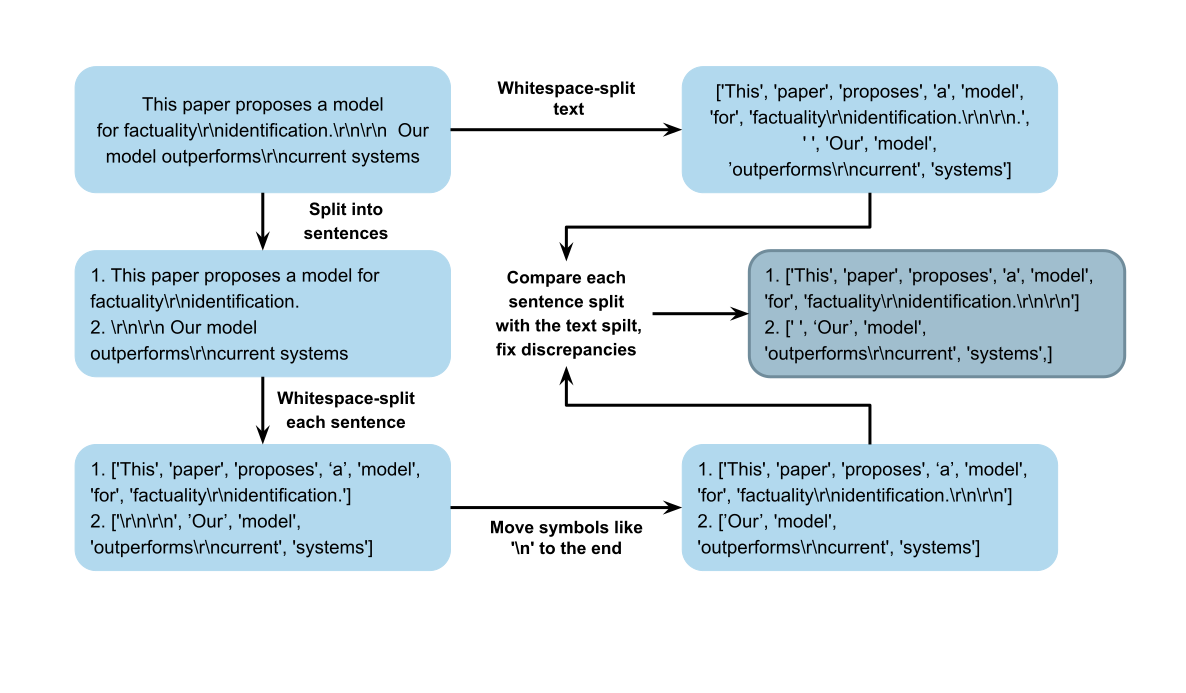}}
\caption{Preprocessing for Augmentation Pipeline}
\label{fig:example}
\end{figure*}

See Figure~\ref{fig:example}.
\label{appendix:pipeline}

\end{document}